\documentclass[11pt]{article}

\usepackage[preprint]{acl}

\usepackage{times}
\usepackage{latexsym}
\usepackage[T1]{fontenc}
\usepackage[utf8]{inputenc}
\usepackage{microtype}
\IfFileExists{inconsolata.sty}{\usepackage{inconsolata}}{}

\usepackage{graphicx}
\usepackage{booktabs}
\usepackage{amsmath,amssymb,amsthm}
\usepackage{bm}
\usepackage{xcolor}
\usepackage{colortbl}
\usepackage{subcaption}
\usepackage{enumitem}
\usepackage{xspace}
\usepackage{algorithm}
\usepackage{algorithmic}
\usepackage{cleveref}   % after hyperref (loaded by acl.sty)
\usepackage{listings}   % two-line code comparison in Section 2.1

\definecolor{cPeek}{HTML}{D55E00}    % identity-peeking commit rules (LLaDA low-confidence, Fast-dLLM threshold, Dream MaskGIT+)
\definecolor{cBlind}{HTML}{0072B2}   % identity-blind commit rules (max-prob / margin / entropy ordering)
\definecolor{cRand}{HTML}{8856A7}    % random order (identity-blind, confidence-free)
\definecolor{cFFS}{HTML}{009E73}     % Fork-First Sampling, faithful mode
\definecolor{cFair}{HTML}{009E73}    % Fork-First Sampling, fair mode
\definecolor{cBase}{HTML}{B07AA1}    % mitigation baselines (instruction, prompt-specified gender, logit bias)
\definecolor{cBelief}{HTML}{1A1A1A}  % one-pass belief (reference quantity; black line in figures)
\definecolor{cBest}{HTML}{E8F4EC}    % table cell highlight

\newtheorem{definition}{Definition}
\newtheorem{lemma}{Lemma}
\newtheorem{proposition}{Proposition}

\newcommand{\ffs}{\textsc{FFS}\xspace}                 % Fork-First Sampling
\newcommand{\ffsfaith}{\textsc{FFS}-faithful\xspace}
\newcommand{\ffsfair}{\textsc{FFS}-fair\xspace}

\newcommand{\llada}{LLaDA\xspace}
\newcommand{\dream}{Dream\xspace}

\newcommand{\M}{\mathcal{M}}                 % masked positions
\newcommand{\G}{\mathcal{G}}                 % group lexicon
\newcommand{\E}{\mathbb{E}}
\newcommand{\1}{\mathbb{1}}
\newcommand{\logit}{\operatorname{logit}}
\newcommand{\belief}{b}                      % one-pass belief at the fork
\newcommand{\share}{s}                       % behavioral share
\newcommand{\target}{\tau}                   % FFS target distribution
\newcommand{\fork}{f}                        % fork position
\newcommand{\Gam}{\Gamma}                    % odds-inflation factor (Lemma 1)

\newcommand{\figorbox}[3]{\IfFileExists{#1}{\includegraphics[width=#2]{#1}}{\fbox{\parbox[c][#3][c]{0.95#2}{\centering\small\textcolor{red}{[figure pending: \detokenize{#1}]}}}}}
\newcommand{\yes}{\textcolor{cPeek}{$\checkmark$}}
\newcommand{\no}{\textcolor{cBlind}{$\times$}}
\newcommand{\dotp}{\textcolor{cPeek}{\textbullet}}
\newcommand{\dotb}{\textcolor{cBlind}{\textbullet}}
\newcommand{\dotr}{\textcolor{cRand}{\textbullet}}
\newcommand{\dotg}{\textcolor{black!70}{\textbullet}}
\newcommand{\dotf}{\textcolor{cFFS}{\rule[0.1ex]{0.5em}{0.5em}}}
\newcommand{\dotfair}{\textcolor{cFair}{\fboxsep=0pt\fboxrule=0.6pt\raisebox{0.1ex}{\fbox{\phantom{\rule{0.4em}{0.4em}}}}}}
\newcommand{\dotbase}{\textcolor{cBase}{\textbullet}}

\title{Who Gets Unmasked? Sampler-Induced Bias\\in Diffusion Language Models}

\author{Xujun Che, Farsheed Haque, Depeng Xu, Xi Niu, Haopeng Zhang\\
University of North Carolina at Charlotte, Charlotte, NC, USA\\
\texttt{\{xche,fhaque,dxu7,xniu2,hzhang55\}@charlotte.edu}
}

\begin{document}
% Full-width teaser under the title block (Figure 1)
\makeatletter
\let\acl@maketitle\@maketitle
\def\@maketitle{\acl@maketitle
  \begin{minipage}{\textwidth}
    \centering
    \includegraphics[width=\textwidth]{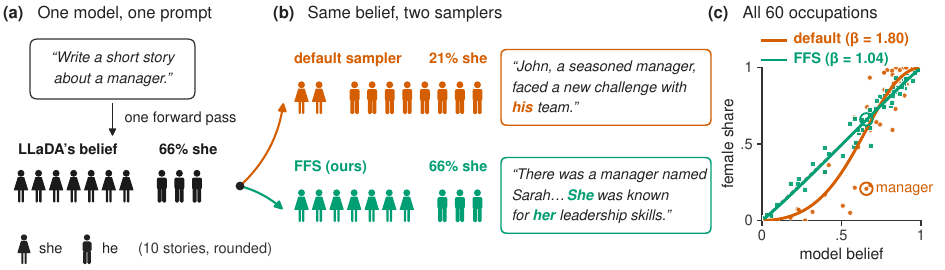}
    \captionof{figure}{(a) \llada's one-pass belief that a story's manager is a woman. (b) The default sampler reverses it; \ffs keeps it. (c) Over 60 occupations, the default sampler amplifies the belief and \ffs follows it.}
    \label{fig:teaser}
  \end{minipage}\vspace{0.4em}}
\makeatother
\maketitle

\begin{abstract}
Diffusion language models (dLLMs) generate by gradually unmasking a canvas, so their samplers decide not only which token to write but also when to commit it.
We show that this second decision changes who appears in generated text, a \emph{sampler-induced bias} on top of the model's own.
Common commit rules, including LLaDA's default, commit the positions whose sampled tokens are most probable; this \emph{peeking} favors majority options.
To separate the sampler from the model, we read the model's belief about a fork, such as a protagonist's gender, in one forward pass, and show that this belief matches exact sampling from the model.
Measured against it, peeking samplers on LLaDA and Dream amplify occupational gender stereotypes and collapse opinion distributions, and confidence-based ordering shifts gender-balanced occupations toward men, more so under faster parallel decoding.
We propose Fork-First Sampling, which decides the fork before anything else is written: it follows the belief or any target distribution, improves fluency, and avoids the stories that remark on the protagonist's gender, which prompt-based fixes produce, at the cost of one forward pass.
Fairness evaluations of dLLMs should report the commit rule.
\end{abstract}

\section{Introduction}
\label{sec:intro}

Diffusion language models (dLLMs) such as LLaDA \citep{nie2025llada} and Dream \citep{ye2025dream} now perform on par with autoregressive (AR) models of similar size.
They generate by unmasking a canvas over several steps, and at each step the sampler makes two decisions: which masked positions to commit, and which tokens to write there.
Representational harms in generated narratives, such as the erasure of minority groups \citep{shieh2025intersectional}, are attributed to the model, and so are the fairness and trustworthiness evaluations of dLLMs \citep{mo2026trustldm}.

Figure~\ref{fig:teaser} shows that this attribution is incomplete. With the model and prompts fixed, LLaDA's default sampler writes a female manager in 21\% of stories although the model itself puts 66\% of its probability on a female protagonist (Figure~\ref{fig:teaser}a, b); across 60 occupations the slope of its female share on the belief is 1.80, where exact sampling would give 1 (Figure~\ref{fig:teaser}c).
For AR models, sampling at $T=1$ draws exactly from the model, so decoding affects representation only through temperature and truncation \citep{dhamala2022analysis}.
dLLMs break this link: LLaDA's default rule samples a token at every masked position and commits the positions whose sampled tokens are most probable, so a sampled minority token is less likely to be kept, and the output at $T=1$ is not a sample from the model; we call such rules \emph{peeking}.
Prior and concurrent work documents the resulting loss of diversity and entropy \citep{hayakawa2025demystifying,tang2026sampler,fang2026locally,zhang2026differences,jeon2026low}; to our knowledge, its effect on who appears in the text has not been studied.

Work on AR models separates the distributions a model can describe from those it samples \citep{meister2025benchmarking,jang2026instruction}. We separate representation in dLLM generation into the model's \emph{belief} about a fork, such as the gender of a protagonist, and the \emph{amplification} that the sampler applies to that belief, which we call sampler-induced bias.
Our contributions are:
\begin{itemize}[leftmargin=*,itemsep=1pt,topsep=2pt]
  \item \textbf{Decomposition and audit.} One forward pass on the fully masked canvas reads the model's belief about a fork; it matches exact sampling from the model at $1/11{,}520$ of the cost of a sampling audit.
  \item \textbf{Two channels of amplification.} In the \emph{token channel}, peeking commit rules, the defaults of LLaDA and Fast-dLLM \citep{wu2025fastdllm} and an option in Dream, multiply the odds of the more probable option by a factor of at least one (Lemma~\ref{lem:tilt}). In the \emph{ordering channel}, committing the most confident positions first writes male-leaning context before the first pronoun, more so when more tokens are committed per step.
  \item \textbf{Fork-First Sampling (\ffs)} draws the fork from the belief (faithful mode) or a target (fair mode) before anything else is written and hands the canvas to any sampler. It costs one forward pass, controls the fork distribution exactly, works with greedy decoding, and, unlike prompt-specified gender, keeps fluency and the story's content. In both models it meets a 50--50 target within 1.4 points and raises fluency.
  \item \textbf{Evidence} on LLaDA-8B, LLaDA-1.5 and Dream-7B across occupational stories, GlobalOpinionQA and TANGO, plus a synthetic model with a known belief.
\end{itemize}

\section{Why Samplers Amplify Majorities}
\label{sec:amplify}

\subsection{Sampling in Masked Diffusion Language Models}
\label{sec:background}

Masked diffusion language models \citep{austin2021structured,sahoo2024simple,shi2024simplified} are trained to reconstruct masked tokens, and the network output at a masked position estimates the distribution of the clean token given the unmasked context \citep{ou2025absorbing}.
Generation starts from a canvas $x$ of $L$ mask tokens following the prompt $c$.
At step $t$, with masked set $\M_t$ and position conditionals $p_i^{(t)}$, the sampler
(i) proposes a token $y_i\sim q_i^{(t)}\propto\big(p_i^{(t)}\big)^{1/T}$ at every masked position, optionally after top-$p$ truncation;
(ii) assigns each position a score $s_i$; and
(iii) selects a commit set $U_t\subseteq\M_t$ (the $k$ highest scores, the scores above a threshold, or a random subset) and writes $y_i$ into $x$ for $i\in U_t$.
Steps (ii) and (iii) form the \emph{commit rule}. Committing the highest-scoring positions first decodes in \emph{confidence order}; drawing $U_t$ uniformly decodes in \emph{random order}.
Confidence-based remasking of this kind comes from Mask-Predict and MaskGIT \citep{ghazvininejad2019mask,chang2022maskgit}.
LLaDA decodes semi-autoregressively in blocks of 32 positions; Dream commits anywhere on the canvas.\footnote{We compute all scores in float64; low-precision Gumbel sampling lowers the effective temperature \citep{zheng2025masked}.}

\begin{table}[t]
\centering
\small
\setlength{\tabcolsep}{3.5pt}
\begin{tabular}{@{}llcl@{}}
\toprule
Rule & Score $s_i$ & Peeks & Used by \\
\midrule
Low-confidence & $p_i(y_i)$ & \yes & LLaDA \\
Threshold & $p_i(y_i)\ge\lambda$ & \yes & Fast-dLLM \\
MaskGIT+ & $q_i(y_i)$ & \yes & Dream (opt.) \\
Soft low-conf.\ & $p_i(y_i)^{1/T'}\!+\!\epsilon$ & \yes & TLC \\
Confidence & $\max_v p_i(v)$ & \no & our patch \\
Entropy & $-H(q_i)$ & \no & Dream (ex.) \\
Margin & $p_i^{(1)}-p_i^{(2)}$ & \no & Dream (opt.) \\
Random & uniform & \no & --- \\
Moment & $\lVert p_i\rVert_\beta^\beta$ & \no & \citeauthor{hayakawa2025demystifying} \\
\bottomrule
\end{tabular}
\caption{Commit rules. \yes: the score peeks at the drawn token.}
\label{tab:rules}
\end{table}

Table~\ref{tab:rules} sorts common commit rules, including the two-temperature sampler TLC \citep{olausson2026tale}, by whether the score depends on the token that was sampled: rules whose score does \emph{peek} at it, and \emph{identity-blind} rules whose score does not (formalized in Definition~\ref{def:peek}).
The two most widely used open dLLMs default to different classes: LLaDA's low-confidence rule peeks, whereas the entropy ordering in Dream's example code does not.
Whether a dLLM amplifies majorities at $T>0$ therefore depends in part on an implementation choice that is rarely reported.
Making LLaDA's rule identity-blind changes one line, and at $T=0$ both lines compute the same score:
\begin{lstlisting}[language=Python,basicstyle=\ttfamily\scriptsize,columns=fullflexible,frame=none,aboveskip=3pt,belowskip=3pt]
s = p.gather(-1, x0[..., None])[..., 0]  # peeking
s = p.max(dim=-1).values                 # blind
\end{lstlisting}

\subsection{Peeking Commit Rules Tilt the Odds}
\label{sec:theory}

\begin{definition}[Peeking]
\label{def:peek}
A commit rule is \emph{identity-blind} if, given the canvas and the rule's own randomness, the commit set $U_t$ is independent of the proposals $\{y_i\}_{i\in\M_t}$. Otherwise the rule \emph{peeks}.
\end{definition}

\begin{figure*}[t]
  \centering
  \figorbox{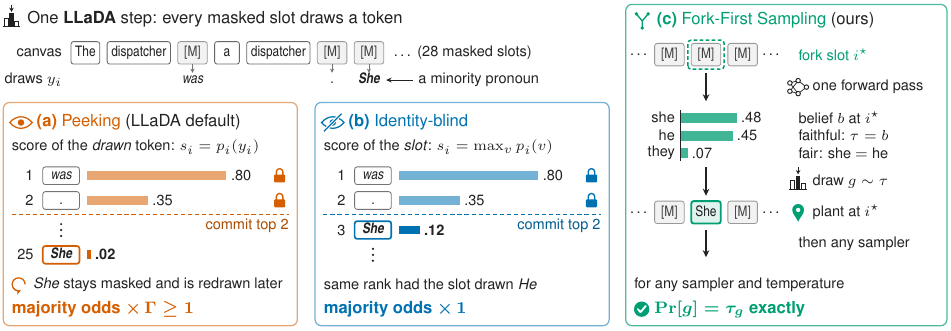}{\textwidth}{4.5cm}
  \caption{One \llada step. Peeking ranks the position that drew \emph{She} 25th of 28; the identity-blind rule ranks it 3rd. \ffs plants the fork token first.}
  \label{fig:method}
\end{figure*}

Identity-blind rules choose positions first and sample tokens afterwards; the order of the two operations does not matter to them.
Figure~\ref{fig:method}a,b shows the difference on one real \llada step: the position that drew the minority pronoun \emph{She} ranks 25th of 28 under peeking, because its score is the probability of the drawn token, and 3rd under the identity-blind rule, whose score would be the same had it drawn \emph{He}.
Consider a fork position $\fork$, such as the first pronoun that refers to the protagonist, and let $a_t(y)=\Pr[\fork\in U_t\mid y_\fork=y]$ be the probability that $\fork$ is committed at step $t$ when its proposal is $y$.

\begin{lemma}[Commit tilt]
\label{lem:tilt}
Suppose $q_\fork$ and $a_t$ do not depend on what has been committed at other positions. Then for any two values $v,w$,
\begin{align}
\frac{\Pr[x_\fork=v]}{\Pr[x_\fork=w]}&=\frac{q_\fork(v)}{q_\fork(w)}\,\Gam_{vw},\label{eq:tilt}\\
\Gam_{vw}&=\frac{\sum_t S_t\,a_t(v)}{\sum_t S_t\,a_t(w)},\nonumber
\end{align}
where $S_t$ is the probability that $\fork$ is still masked at step $t$.
For identity-blind rules, $\Gam_{vw}=1$.
If the rule peeks and $a_t$ is nondecreasing in a score that is nondecreasing in $p_\fork(y)$, then $p_\fork(v)\ge p_\fork(w)$ implies $\Gam_{vw}\ge1$.
\end{lemma}

Without peeking the odds are $(p_\fork(v)/p_\fork(w))^{1/T}$, set by the temperature; peeking multiplies them by $\Gam$ (proofs of all results are in Appendix~\ref{app:proofs}). For scale, the $\Gam=2.3$ we measure for \llada turns a 40\% option into a 22\% one.
Unlike the implicit temperature of \citet{hayakawa2025demystifying}, Lemma~\ref{lem:tilt} needs no large-$N$ limit and covers threshold rules and forced commits ($a_t\equiv1$); unlike entropy bounds \citep{fang2026locally,zhang2026differences,jeon2026low}, it gives the odds between the two values that decide which group is written.
$\Gam$ is observable as the ratio of the commit rates of \emph{majority} and \emph{minority} proposals, that is, proposals from the group with more and with less probability mass at their position.

\begin{proposition}[Identity-blind rules are exact]
\label{prop:blind}
If the rule is identity-blind, then conditional on $\fork$ being committed at step $t$, $x_\fork\sim q_\fork^{(t)}$. If in addition $T=1$, one token is committed per step, and the position conditionals are consistent with a joint distribution $P$, then every adaptive order samples exactly from $P$.
\end{proposition}

The second part is Proposition~3 of \citet{hayakawa2025demystifying}.
Its contrapositive is a test: if two identity-blind orders at $T=1$ with one token per step disagree on the fork, the conditionals are inconsistent, as is common for masked language models \citep{young2023inconsistencies,hennigen2023deriving,yamakoshi2022probing,kim2026path}. In our data they agree, which gives an operational meaning to the belief that we read from the model's conditionals in the next section.

\paragraph{When order matters.}
At $T\neq1$ stepwise commits do not compose into one joint distribution: early commits receive sharpened marginals and late ones sharpened conditionals, so the order changes the output even for a consistent model.
In a consistent synthetic model, confidence order favors the majority more than random order at $T\le0.5$ but not at $T=1$ (Appendix~\ref{app:sim}).
This is the theoretical basis of the ordering channel, in which confidence order shapes the context written before the fork; its male direction in real models is a property of the models.

\section{Auditing and Controlling the Fork}
\label{sec:method}

\subsection{Reading the Belief in One Forward Pass}
\label{sec:audit}

By linearity, the expected number of tokens from a group lexicon $\G$ (e.g., she/her/hers/herself) under a joint distribution $P$ is $\E_P[N_\G]=\sum_iP(x_i\in\G\mid c)$, and one forward pass on the fully masked canvas returns every term, since $p_i^{(0)}$ estimates $P(x_i\mid c)$.
For the female and male lexicons $\G_F$ and $\G_M$, the \emph{count belief} $\sum_ip^{(0)}_i(\G_F)/\sum_i[p^{(0)}_i(\G_F)+p^{(0)}_i(\G_M)]$ is compared with pronoun counts in sampled text.
The \emph{fork belief} $\belief$ is the same ratio at the position $i^\star$ of the first block with the largest pronoun mass: the belief about the first gendered reference, which we compare with the share of female protagonists. The two differ when stories about one gender use more pronouns, so neither substitutes for the other.
The audit costs one forward pass, against 11,520 for a sampling audit with 120 stories of 96 steps.\footnote{Dream is initialized from an AR model and predicts position $i$ from the output at $i-1$; we shift its logits accordingly.}

\subsection{Fork-First Sampling}
\label{sec:ffs}

\begin{algorithm}[t]
\caption{Fork-First Sampling}
\label{alg:ffs}
\small
\begin{algorithmic}[1]
\REQUIRE prompt $c$, groups $\G_g$ with nominative forms $\mathcal{N}_g\subseteq\G_g$, mode, sampler $\mathcal{S}$, first-block length $B$
\STATE $p^{(0)}\gets$ model$(c,[\textsc{m}]^L)$ \hfill\COMMENT{one forward pass}
\IF{$\sum_{i,g}p^{(0)}_i(\G_g)<0.5$}
  \RETURN $\mathcal{S}(c)$ \hfill\COMMENT{no fork expected}
\ENDIF
\STATE $i^\star\gets\arg\max_{i\le B}\sum_g p^{(0)}_i(\G_g)$; \ $\belief_g\propto p^{(0)}_{i^\star}(\G_g)$
\STATE $\target\gets\belief$ (faithful) or $\target_{\text{they}}=\belief_{\text{they}},\ \target_F=\target_M=\tfrac{1-\belief_{\text{they}}}{2}$ (fair)
\STATE $j^\star\gets\arg\max_{i\le B}\sum_g p^{(0)}_i(\mathcal{N}_g)$ \hfill\COMMENT{subject slot}
\STATE draw $g\sim\target$; \ $x_{j^\star}\gets\arg\max_{v\in\mathcal{N}_g}p^{(0)}_{j^\star}(v)$
\RETURN $\mathcal{S}(c,x)$ \hfill\COMMENT{any sampler, any $T$}
\end{algorithmic}
\end{algorithm}

\ffs (Algorithm~\ref{alg:ffs}, Figure~\ref{fig:method}c) decides the fork before anything else is written: it reads the belief at $i^\star$, draws a group $g$ from a target $\target$ ($\target=\belief$ in faithful mode; in fair mode \emph{they} keeps its share and \emph{she} and \emph{he} split the rest), plants the group's most probable nominative form (\emph{she}, \emph{he}, \emph{they}, in the model's preferred case and spacing) at $j^\star$, the first-block position with the largest nominative-pronoun mass, so that the planted pronoun is the protagonist's subject, and runs an unmodified sampler on the planted canvas. If fewer than half a gendered token is expected on the canvas, it does not intervene.

\begin{proposition}[Exact fork control]
\label{prop:ffs}
When \ffs intervenes, $\Pr[x_{j^\star}\in\G_g]=\target_g$ for every $g$, for any sampler and temperature.
\end{proposition}

The planted position and token depend only on the step-0 distribution and $g$.
\ffs controls the fork's marginal; the rest of the text follows the sampler conditioned on the plant, and whether later references agree is measured as \emph{protagonist consistency}.
\ffs costs one forward pass (2\% at 48 steps), leaves the input unchanged, composes with any sampler, and controls the distribution even under greedy decoding, where the only randomness is the draw of $g$.
Unlike a logit bias \citep{an2026dlmswai}, which shifts group tokens at every step and thereby the order, \ffs commits once, at the position of a specified variable, with a value drawn from a stated distribution.

\section{Experiments}
\label{sec:experiments}
\label{sec:results}

\subsection{Setup}
\label{sec:setup}

\paragraph{Models and tasks.}
LLaDA-8B-Instruct \citep{nie2025llada} and Dream-v0-Instruct-7B \citep{ye2025dream}, with LLaDA-1.5 \citep{zhu2025llada15} as a replication; Qwen2.5-7B-Instruct \citep{qwen2024qwen25}, for which sampling at $T=1$ is exact, is an AR reference.
\emph{Occupational stories}: the 60 Winogender occupations \citep{rudinger2018gender} with their female shares in U.S.\ Bureau of Labor Statistics (BLS) data \citep{kirk2021bias,lucy2021gender}, three templates (180 prompts), 20 stories per prompt and sampler. Qwen2.5-32B-Instruct \citep{zheng2023judging} labels the protagonist's gender (female, male, they, unclear, multiple) and rates fluency; we check the labels against pronoun-counting rules, a second judge \citep[Llama-3.3-70B;][]{grattafiori2024llama3} and human annotation of 300 stories ($\kappa=0.90$; judge accuracy 98.5\%; Appendix~\ref{app:labels}).
\emph{Opinions}: 250 GlobalOpinionQA questions \citep{durmus2024towards}, 20 answers each; the belief is the answer-letter distribution at the answer position of the fully masked canvas and the behavior is the first letter of the generated answer; the two can differ \citep{wang2024my}.
\emph{Misgendering}: 600 TANGO prompts \citep{ovalle2023tango}, 100 per declared family (she, he, they, xe, ey, fae), 4 continuations each.

\paragraph{Samplers and decoding.}
Peeking (LLaDA's low-confidence rule; Dream's MaskGIT+), identity-blind (the one-line patch of Appendix~\ref{app:impl}; for \dream also entropy ordering), random order, greedy decoding, \ffsfaith and \ffsfair, the latter also with greedy decoding, and the mitigation baselines of Table~\ref{tab:methods}.
Stories have 96 tokens decoded in 48 steps (two per step); LLaDA uses blocks of 32 and \dream none.
We decode at $T=0.5$ with top-$p=0.9$ \citep{holtzman2020curious}, the highest-entropy setting at which identity-blind samplers write fluent English; \dream defers end-of-sequence commits to the last step (Appendix~\ref{app:details}).

\paragraph{Metrics and statistics.}
We fit the binomial GLM $\logit\share=\alpha+\beta\logit\belief$ across prompts, where $\share$ is the share of female protagonists: $\beta=1$ is faithful to the belief, $\beta>1$ amplifies the majority, and $\alpha<0$ shifts toward men. Prompts with $0.4\le\belief\le0.6$ are \emph{belief-balanced}.
For \ffs we report the target error $|\text{pooled female share}-\tfrac12|$ and protagonist consistency, the share of stories whose protagonist has the planted gender. Quality is judged fluency (1--5), the rate of \emph{marked} stories that comment on the protagonist's gender as unusual, distinguishability from the default sampler's stories (AUC), and distinct-2 and self-BLEU-4 within a prompt \citep{shaib2024standardizing}.
For opinions we report the ratio of behavior to belief entropy, the total variation (TV) between them, and the share of collapsed questions (top answer $\ge0.9$ in behavior but $\le0.7$ in the belief), each against the floor of sampling exactly from the belief; the misgendering rate is the mean share of a continuation's third-person pronouns outside the declared family.
Confidence intervals are 95\% cluster bootstraps over occupations (at least 1,000 resamples), and samplers are compared on the same prompts and seeds.

% Subsection order mirrors the argument (../OUTLINE.md §7): the belief is a valid reference ->
% samplers amplify it -> temperature cannot undo it -> where it enters -> FFS vs baselines -> other forks.

\begin{figure*}[t]
  \centering
  \figorbox{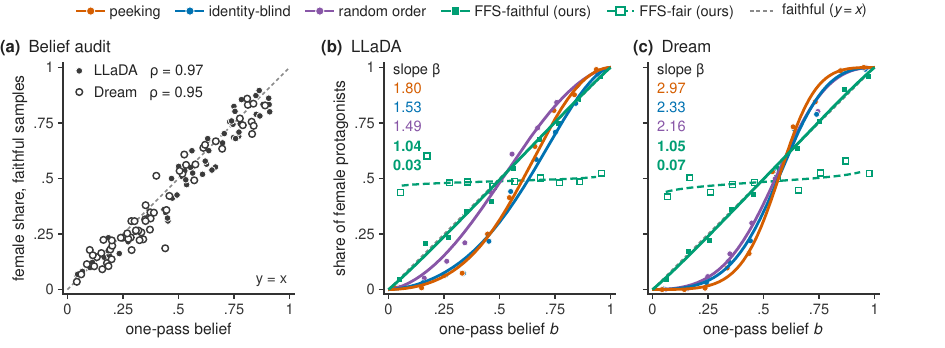}{\textwidth}{5.2cm}
  \caption{(a) Belief against faithful samples. (b, c) Belief against sampler behavior, with GLM slopes $\beta$ (1: faithful).}
  \label{fig:scurves}
\end{figure*}

\subsection{The One-Pass Belief Predicts Faithful Behavior}
\label{sec:res-audit}

We audit the one-pass belief against faithful samples: untruncated $T=1$, one token per step, and random and confidence orders, which are exact for a consistent model (Proposition~\ref{prop:blind}), with 120 stories for each of the 60 prompts of one template.
The belief tracks the samples closely (Figure~\ref{fig:scurves}a; Spearman $\rho=0.97$ for \llada and $0.95$ for \dream), with errors close to those expected from sampling noise alone and a small female offset that lies inside a $\pm5$-point equivalence margin.
The two orders agree on the fork distribution, so the test after Proposition~\ref{prop:blind} detects no inconsistency (Appendix~\ref{app:details}).
The belief already encodes the occupational stereotype (Spearman correlation with BLS female shares 0.80 and 0.76); what follows measures what the sampler adds to it.

\subsection{Peeking Samplers Amplify the Majority}
\label{sec:res-amp}

Peeking bends the belief--behavior curve beyond what the same ordering does without it (Figure~\ref{fig:scurves}b, c).
Its slope exceeds that of the identity-blind rule in both models, by $\Delta\beta=0.28$ $[0.08, 0.51]$ for \llada and $0.64$ $[0.26, 1.29]$ for \dream (slopes in Table~\ref{tab:methods}).
The gap is not an artifact of labels or fluency: it holds under a second judge, shrinks but keeps its sign under rule-based labels, which human annotators find correct on only 23\% of disputed stories (the judge: 72\%), and persists with one token per step, where both rules write equally fluent text, and at every level of parallelism we tested (Appendices~\ref{app:labels} and~\ref{app:amp}).
Peeking also narrows the text itself: at matched fluency it lowers distinct-2 by 0.085.
LLaDA-1.5 replicates the gap ($\Delta\beta=0.23$ $[0.08, 0.44]$) and the male shift under confidence ordering (Appendix~\ref{app:amp}).

Confidence ordering adds a second, male-leaning shift.
On belief-balanced prompts, \llada's peeking and identity-blind rules write about 14 points fewer female protagonists than parity, whereas random order does not.
In \dream every sampler except \ffs shifts toward men, random order included, and both \ffs modes show no shift in either model ($|\alpha|\le0.07$).
The shift is relative to each model's own belief; LLM-written stories often feature more women overall \citep{chen2025more}.

\begin{figure}[t]
  \centering
  \figorbox{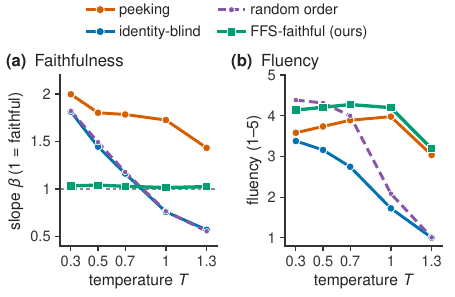}{\columnwidth}{5cm}
  \caption{\llada over temperature (top-$p=1$): (a) slope on the belief, (b) fluency. Only \ffsfaith is faithful and fluent at every temperature.}
  \label{fig:temp}
\end{figure}

\subsection{Temperature Cannot Undo Peeking}
\label{sec:res-temp}

Raising the temperature, the one setting users control, barely changes a peeking sampler (Figure~\ref{fig:temp}).
As the temperature rises from 0.3 to 1.3, \llada's identity-blind slope falls roughly as $1/T$ and crosses 1, while the peeking slope stays above 1.4 (Figure~\ref{fig:temp}a): at every temperature, peeking is at least about as steep as identity-blind decoding at $T=0.5$.
Lemma~\ref{lem:tilt} explains why: as $T$ rises the identity-blind rule commits more and more tokens other than the position's most probable one, whereas the peeking rule keeps rejecting them, so the gap between the two rules grows with the temperature (Appendix~\ref{app:amp}).

Reaching the belief through temperature therefore costs fluency: identity-blind samplers reach a slope of 1 only between $T=0.7$ and $T=1$, where their text degrades (at $T=1$, fluency 1.72 and CJK characters in 87\% of outputs).
\ffs removes this trade-off because it draws the fork from the untempered belief: \ffsfaith stays faithful at every temperature and is more fluent than the peeking sampler throughout (Figure~\ref{fig:temp}).
For the AR reference Qwen, where sampling at $T=1$ is exact, greedy decoding raises the female share only from 0.786 to 0.817.

\begin{figure*}[t]
  \centering
  \figorbox{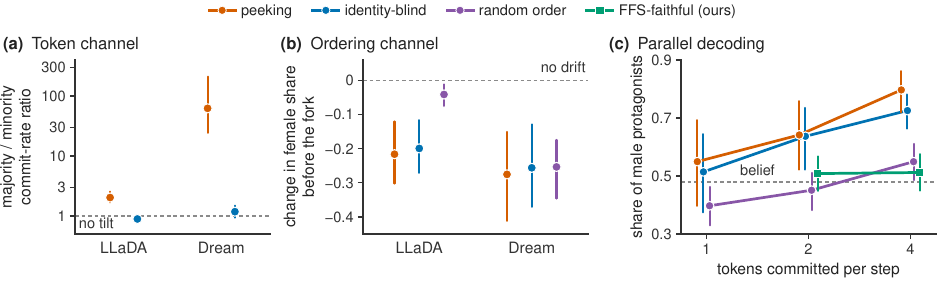}{\textwidth}{4.5cm}
  \caption{(a) Majority/minority commit-rate ratio. (b) Pre-fork drift of the fork position's female share. (c) Male share against tokens per step. Belief-balanced prompts in (b, c).}
  \label{fig:mech}
\end{figure*}

\begin{table*}[t]
\centering
\small
\setlength{\tabcolsep}{3.6pt}
\begin{tabular}{@{}lcc rrrrr rrrrr@{}}
\toprule
 & & & \multicolumn{5}{c}{\llada} & \multicolumn{5}{c}{\dream} \\
\cmidrule(lr){4-8}\cmidrule(l){9-13}
Method & \shortstack{Edits\\prompt} & \shortstack{Extra\\passes} & $\beta$ & TE & $\Delta$Flu. & Mark. & AUC & $\beta$ & TE & $\Delta$Flu. & Mark. & AUC \\
\midrule
\multicolumn{13}{@{}l}{\emph{Commit rules}} \\
\dotg\ Greedy & & & 2.17 & .123 & $-0.20$ & 0.0 & -- & 3.48 & .142 & $-0.45$ & 0.0 & -- \\
\dotp\ Peeking (default) & & & 1.80 & .116 & 0 & 0.3 & -- & 2.97 & .068 & 0 & 0.3 & -- \\
\dotb\ Identity-blind & & & 1.53 & .087 & $-0.59$ & 0.6 & .68 & 2.33 & .062 & $-0.16$ & 0.3 & .64 \\
\dotr\ Random order & & & 1.49 & .180 & $+0.56$ & 0.6 & .86 & 2.16 & .048 & $-0.15$ & 0.5 & .93 \\
\midrule
\multicolumn{13}{@{}l}{\emph{Mitigations on the default sampler}} \\
\dotbase\ Instruction & \checkmark & & 1.28 & .106 & $-0.03$ & 0.2 & .83 & 0.97 & .032 & $-0.17$ & 0.6 & .69 \\
\dotbase\ Prompt gender, balanced & \checkmark & & 0.01 & .004 & $-0.55$ & 3.4 & .83 & 0.02 & .006 & $-0.20$ & 2.9 & .77 \\
\dotbase\ Prompt gender, belief-matched & \checkmark & & 1.02 & -- & $-0.54$ & 3.4 & .84 & 1.00 & -- & $-0.17$ & 1.8 & .78 \\
\dotbase\ One-shot logit bias & & +1 & 0.85 & .046 & $-0.09$ & 0.2 & .60 & 0.67 & .083 & $-0.15$ & 0.2 & .58 \\
\dotp\ $T=1$ & & & 1.73 & .165 & $+0.22$ & 0.4 & .68 & 1.87 & .057 & $-0.01$ & 0.3 & .73 \\
\dotp\ TLC & & & 1.88 & .187 & $+0.78$ & 0.3 & .76 & 2.54 & .043 & $+0.11$ & 0.3 & .88 \\
\midrule
\multicolumn{13}{@{}l}{\emph{Fork-first}} \\
\dotf\ \ffsfaith & & +1 & 1.04 & -- & $+0.45$ & 0.3 & .73 & 1.05 & -- & $+0.30$ & 0.2 & .73 \\
\dotfair\ \ffsfair & & +1 & 0.03 & .006 & $+0.42$ & 0.3 & .72 & 0.07 & .014 & $+0.28$ & 0.4 & .71 \\
\dotfair\ \ffsfair + greedy & & +1 & 0.02 & .004 & $+0.27$ & 0.0 & -- & 0.05 & .024 & $-0.10$ & 0.7 & -- \\
\bottomrule
\end{tabular}
\caption{Methods in the default setting. TE: target error; Mark.: \% marked narratives; AUC: separability from default outputs (0.5 is best); --: not applicable.}
\label{tab:methods}
\end{table*}

\subsection{Where the Amplification Enters}
\label{sec:res-mech}

\paragraph{Token channel.}
Before the fork, peeking commits majority proposals far more readily than minority proposals of the same confidence, and the identity-blind rule does not (Figure~\ref{fig:mech}a).
The implied tilt $\Gam$ of Lemma~\ref{lem:tilt} is about 2.3 for \llada and about 50 for \dream, whose peeking score uses the sharper tempered distribution (Appendix~\ref{app:mech}).

\paragraph{Ordering channel.}
The protagonist's gender is settled before the first pronoun is written: the first gendered word almost always determines it, and under confidence ordering the first pronoun's position is usually already decided when it is committed.
On belief-balanced prompts, confidence ordering moves that position toward male before the fork, and random order barely does; \dream drifts under every order, which is why its shift does not depend on the order (Figure~\ref{fig:mech}b).
Parallel decoding strengthens the drift: under \llada's peeking rule the male share on belief-balanced prompts rises from 0.55 with one token per step to 0.80 with four (Figure~\ref{fig:mech}c), and in \dream from 0.67 with two to 0.80 with four, so faster decoding hardens the male default.
A context transplant shows that the context written before the fork carries the shift, in line with the propagation of early commits \citep{kim2026early}: finished in the same random order, canvases written in confidence order yield 69\% male protagonists on belief-balanced prompts and those written in random order 49\% (+19 points $[7, 29]$; Appendix~\ref{app:mech}).
\ffs decides the fork before any context exists and so bypasses both channels: with four tokens per step its male share on belief-balanced prompts stays at 0.51--0.52 (peeking: 0.80).

\subsection{Fork-First Sampling versus Mitigation Baselines}
\label{sec:res-ffs}

\ffs is the only method that meets the target without lowering fluency, adding marked narratives, or changing what the story is about (Table~\ref{tab:methods}).
\ffsfair nearly eliminates the target error, keeps the planted protagonist in 99\% and 98\% of stories, and raises fluency over the default sampler, as human ratings confirm.
Specifying the gender in the prompt also meets the target, but it lowers fluency and raises the rate of marked narratives from under 0.5\% to about 3\%, echoing the marked personas of \citet{cheng2023marked}.
It also changes what the story is about: a classifier separates prompt-specified stories from the default sampler's stories with the same protagonist gender more easily than it separates \ffs stories (Table~\ref{tab:methods}; Appendix~\ref{app:ffs}), even when the prompted gender is drawn from the belief.
\ffs changes whom the story is about; the prompt also changes what it is about.

The other baselines fall short.
The one-line identity-blind patch removes only the token channel: its slopes stay at 1.53 and 2.33, with lower fluency.
An instruction not to infer gender from the occupation \citep[cf.][]{ganguli2023capacity} lowers $\beta$ mainly by leaving the gender unstated, in about half of the stories; a one-shot logit bias that balances the belief works only partially; higher temperature, random order and TLC leave $\beta$ well above 1.
\ffs also controls the fork under greedy decoding, lowering the target error from 0.12 and 0.14 to 0.004 and 0.024. Its distinct-2 matches the default sampler's (within 0.011); self-BLEU-4 rises by 0.02--0.07.
It stays at the belief with four tokens per step and under Fast-dLLM's threshold rule ($\beta=1.02$--$1.06$), and planting the subject form matters: a pronoun of any case, often \emph{her}, more often binds to a secondary character (Appendix~\ref{app:ffs}).

\subsection{Beyond Occupational Gender}
\label{sec:res-beyond}

\paragraph{Opinions.}
At untruncated $T=1$ peeking compresses answer distributions to less than half of the belief's entropy and collapses 14--16\% of questions onto a single answer, far more than the identity-blind rule (Table~\ref{tab:beyond} in Appendix~\ref{app:beyond}).
\ffsfaith, which plants an answer letter drawn from the belief, is as close to the belief as exact sampling from it.
At $T=0.5$ the lower temperature itself collapses the answers, and peeking adds less.

\paragraph{Misgendering.}
Peeking does not increase misgendering on TANGO: the declared pronoun is usually the majority proposal at the first reference, which Lemma~\ref{lem:tilt} predicts peeking favors.
Neopronouns, however, are misgendered under every sampler, and the one-pass belief predicts this family by family (e.g., \llada misgenders xe in 96\% of cases at the belief level and 95\% when sampling; Appendix~\ref{app:beyond}).
The exception is \dream's xe, whose capitalized form is two tokens that \dream often copies from the prompt.
Misgendering is thus a belief-level problem that no sampler fixes and that one forward pass reveals.
Planting the declared pronoun at the first reference, as \ffs does, lowers misgendering of every neopronoun family in both models (e.g., \llada xe from 0.95 to 0.58, \dream ey from 0.72 to 0.38) without raising it for \emph{he} or \emph{they} (Appendix~\ref{app:beyond}).

\section{Related Work}
\label{sec:related}

\paragraph{Decoding in masked diffusion LMs.}
Adaptive orders and commit rules were designed for quality and speed \citep{kim2025train,wu2025fastdllm}.
Samplers can also depart from the model: sample-then-choose acts as an implicit temperature \citep{hayakawa2025demystifying}, and few-step sampling is unfaithful even with an exact denoiser \citep{tang2026sampler}.
Confidence-based remasking lowers diversity and entropy \citep{fang2026locally,zhang2026differences,jeon2026low}, confidence order postpones high-entropy forks and collapses them \citep{olausson2026tale,ni2026flexibility}, and early commits propagate through the output \citep{kim2026early}.
We measure what these effects do to social groups against the model's belief, and identify an ordering channel that does not depend on peeking.

\paragraph{Social bias in generation and where it comes from.}
Generated stories exhibit representational harms around occupation and gender \citep{lucy2021gender,kirk2021bias,cheng2023marked,shieh2025intersectional}, and language models misgender people who use neopronouns \citep{ovalle2023tango,hossain2023misgendered}; decoding is one of their sources \citep{sheng2021societal,dhamala2022analysis}.
Bias amplification is usually measured relative to training data \citep{zhao2017men}, a reference that can itself create apparent amplification \citep{seshadri2024bias}, and bias statistics stay stable only when outputs behave like samples from the learned distribution \citep{taori2023data}.
The distributions a model reports and the ones it generates also differ for opinions \citep{santurkar2023whose,meister2025benchmarking,wang2024my,jang2026instruction} and for misgendering \citep{subramonian2025agree}.
Bias studies of diffusion models of text are at the model level \citep{mo2026trustldm,atabuzzaman2026reliability}, and the same weights refuse differently under diffusion sampling \citep{rahimi2026stepwise}.
We measure what the dLLM sampler adds to the model's own belief.

\paragraph{Controlling who appears.}
At the prompt level, instructions reduce stereotyping \citep{ganguli2023capacity} and specified attributes introduce marked narratives \citep{cheng2023marked}.
At the decoding and model level, dLLM distributions can be steered with stepwise logit biases \citep{an2026dlmswai} or guidance \citep{nisonoff2024unlocking,schiff2024simple}, target distributions can be imposed by training \citep{khalifa2021distributional,jiang2026controlling}, attribute tags can be drawn before writing \citep{garridomunoz2025tagfirst}, and structural anchors can be planted in a dLLM canvas before infilling \citep{lee2026unlocking}.
In image diffusion, samplers change bias amplification \citep{roos2025how}, sampling favors high-density majorities \citep{um2024dont}, and balanced output distributions can hide entangled concepts \citep{chakraborty2025biasmap}.
\ffs needs no training, plants a single token drawn from the model's belief or a stated target, composes with any sampler, and guarantees the fork distribution.

\section{Conclusion}
\label{sec:conclusion}

Who appears in dLLM output depends on the sampler as well as the model.
Peeking amplifies the model's own stereotypes, and confidence ordering adds a male default that faster decoding strengthens.
We offer three recommendations.
Fairness evaluations of dLLMs should report the commit rule (and whether it peeks), the ordering, the number of tokens committed per step, and the temperature.
They should read the model's belief with one forward pass and report the model's bias and the sampler's amplification separately.
When a sensitive fork must be controlled, deciding it first is preferable to editing the prompt or raising the temperature.

% Unnumbered, after the conclusion and before the references; not counted toward the page limit.
\section*{Limitations}

Our main analyses treat gender as binary; \emph{they} and neopronouns are evaluated only on TANGO, and the erasure of non-binary identities in language technology \citep{dev2021harms} is broader than what a pronoun lexicon captures.
All tasks are in English, and the occupations and BLS statistics are U.S.-centric.
Forks are defined by lexicons of single tokens (pronouns and answer letters); forks spread over several tokens, such as names, are not covered.
We study two dLLM families, LLaDA (with LLaDA-1.5) and Dream.
Protagonist labels come from an LLM judge, validated against rules, a second judge and human annotation of a subsample.
The context transplant that attributes the ordering channel to the context was run on LLaDA only.
Opinion beliefs are read from the fully masked marginal at the answer position, and we do not control for option order or labels, which affect survey answers of LLMs \citep{dominguezolmedo2024questioning}.
\ffs controls the first reference; later references follow from the model, and at least 98\% of stories keep the planted gender.
The fair target is a normative choice.

\section*{Ethical Considerations}

\ffs can impose any distribution, including skewed ones; we recommend that deployments disclose their target. Planting a token in an intermediate state can also be used to bypass safety alignment \citep{yamabe2026toward}.
The belief is a descriptive quantity, not a normative standard: faithful mode reproduces what the model believes, fair mode imposes a target, and which is appropriate depends on the application and the harm at stake \citep{blodgett2020language}.
We use no personal data; human annotation was performed by the authors.
All models and datasets are used under their licenses (Appendix~\ref{app:details}).

\bibliography{refs}

\appendix

% Appendices follow the references; reviewers are not required to read them. Plan: ../OUTLINE.md §11.

\section{Proofs}
\label{app:proofs}

\paragraph{Lemma~\ref{lem:tilt}.}
Under the fixed-context assumption, at every step at which $\fork$ is masked the sampler draws $y\sim q_\fork$ and commits it with probability $a_t(y)$, where the probability is over the competitors' scores and the rule's own randomness, neither of which depends on $y_\fork$. Let $S_t$ be the probability that $\fork$ is masked at step $t$; it satisfies $S_{t+1}=S_t\,(1-\sum_y q_\fork(y)a_t(y))$ and does not depend on the value eventually written. Hence $\Pr[x_\fork=v]=q_\fork(v)\sum_t S_t a_t(v)$, and the ratio for $v$ and $w$ is Eq.~\eqref{eq:tilt}. If the rule is identity-blind, $a_t(y)$ does not depend on $y$ and $\Gam_{vw}=1$. If the rule peeks with a score nondecreasing in $p_\fork(y)$ and an acceptance probability nondecreasing in the score, then $p_\fork(v)\ge p_\fork(w)$ implies $a_t(v)\ge a_t(w)$ for every $t$, so $\Gam_{vw}\ge1$. Top-$k$ selection, thresholds with a fallback to the highest score, and forced commits at the end of a block ($a_t\equiv1$) all satisfy these conditions.

\paragraph{Proposition~\ref{prop:blind}.}
If $U_t$ is independent of the proposals given the state, conditioning on $\fork\in U_t$ does not change the distribution of $y_\fork$, so $x_\fork=y_\fork\sim q_\fork^{(t)}$. With $T=1$ and one commit per step, each step selects one position independently of its value and draws the value from $P(x_i\mid c,x_{\text{obs}})$; by the chain rule, the product of these conditionals along any order is $P$ \citep[Prop.~3]{hayakawa2025demystifying}.

\paragraph{Order matters at $T\neq1$.}
Take two binary positions with the consistent joint distribution $P(1,1)=0.35$, $P(1,0)=0.25$, $P(0,1)=0$, $P(0,0)=0.40$, and let $T\to0$. Committing position~1 first gives $x_1=\arg\max P(x_1)=1$ ($P(x_1=1)=0.60$) and then $x_2=1$ ($P(x_2=1\mid x_1=1)=0.58$). Committing position~2 first gives $x_2=0$ ($P(x_2=0)=0.65$) and then $x_1=0$ ($P(x_1=0\mid x_2=0)=0.62$). The two orders yield opposite values of $x_1$, although both are exact at $T=1$: the position committed first receives a sharpened marginal, the other a sharpened conditional.

\paragraph{One-pass belief.}
$N_\G=\sum_i\1[x_i\in\G]$, so $\E_P[N_\G]=\sum_iP(x_i\in\G\mid c)$ by linearity. On the fully masked canvas the network's output at position $i$ estimates $P(x_i\mid c)$, and one forward pass returns all positions.

\paragraph{Proposition~\ref{prop:ffs}.}
$j^\star$ and the token planted for each group are functions of $p^{(0)}$, which depends only on $c$; $g$ is drawn from $\target$ independently of everything else, and the planted token belongs to $\G_g$. None of the samplers we use remask committed tokens, so $x_{j^\star}\in\G_g$ with probability $\target_g$.

\section{Sampler Implementations}
\label{app:impl}

\paragraph{LLaDA.} The official \texttt{generate} draws $y_i=\arg\max_v(\log p_i(v)+T\,\epsilon_{iv})$ with Gumbel noise $\epsilon$ (sampling from $q_i\propto p_i^{1/T}$) and, for \texttt{low\_confidence} remasking, scores $s_i=p_i(y_i)$ with the untempered softmax; within each block of 32 positions it commits the $k$ highest scores per step, where $k$ spreads the block's positions evenly over its steps. \texttt{random} remasking uses uniform scores. Our identity-blind rule replaces $p_i(y_i)$ by $\max_vp_i(v)$ (Section~\ref{sec:background}). Official LLaDA computes scores in bfloat16, which produces ties; we compute them in float64. When $T>1$ we also exclude the mask token from proposals, because tempering otherwise lets it be proposed and committed.

\paragraph{Fast-dLLM.} The threshold rule commits every position with $p_i(y_i)\ge\lambda$ and at least the highest-scoring one \citep{wu2025fastdllm}; its identity-blind variant thresholds $\max_vp_i(v)$.

\paragraph{TLC.} Positions are drawn without replacement with probabilities proportional to $s_i^{1/T'}$ (Gumbel-top-$k$ on $\log s_i/T'$), where $s_i$ is the peeking score, $p_i(y_i)$ for \llada and $q_i(y_i)$ for \dream; we use $T'=1$.

\paragraph{Dream.} \texttt{generation\_utils} divides the logits by $T$, applies top-$p$, and computes $q_i$. \texttt{maskgit\_plus} scores $q_i(y_i)$ (peeking, on the tempered and truncated distribution); \texttt{topk\_margin} scores the gap between the two largest probabilities; \texttt{entropy} scores $-H(q_i)$; \texttt{origin} unmasks each masked position independently with probability $1-s/t$, a random order. The number of positions committed at a step is $\lfloor n(1-s/t)\rfloor$ for $n$ masked positions and the last step commits the rest. Our identity-blind MaskGIT+ scores $\max_vq_i(v)$.

\paragraph{Randomness.} Every sample has its own random streams for proposals and positions, seeded by prompt and sample index, so that samplers are compared on the same random numbers.

\section{Known-Truth Simulation}
\label{app:sim}

\paragraph{Model.} A latent group $g\in\{F,M,\textit{they}\}$ is drawn from a prior whose female share among $\{F,M\}$ is one of 19 values $r\in\{0.05,\dots,0.95\}$; given $g$, each of 48 positions emits a token independently from an emission distribution of its type (four pronoun slots, eight anchor slots, neutral slots). Because all conditionals derive from one joint distribution, they are consistent, and the exact conditional at any state is available in closed form. The first pronoun slot is the fork; its belief is $r$. In the calibrated variant the neutral positions' confidence distribution matches that measured on LLaDA, so that the competition a fork faces is realistic. Each prior gets 2,000 samples per sampler (20,000 for exactness tests); decoding uses blocks of 16 and $T=1$.

\paragraph{Results.} Table~\ref{tab:sim} lists slopes. Identity-blind rules, random order and \ffsfaith recover the belief ($\beta\approx1$); \ffsfair is flat; peeking rules amplify. Chi-square tests of exactness pass for one-token-per-step random and confidence orders and for both \ffs modes (all $p\ge0.09$). Unlike the real models, the synthetic model's peeking slope falls with the number of tokens per step, because with more commits per step a minority proposal at the fork is less often outcompeted; in LLaDA the peeking premium does not depend on $k$ (Section~\ref{sec:res-amp}).

\begin{table}[h]
\centering\small
\begin{tabular}{@{}lccc@{}}
\toprule
Sampler & $k=1$ & $k=2$ & $k=4$ \\
\midrule
\dotp\ Peeking (low-confidence) & 3.34 & 2.81 & 1.12 \\
\dotb\ Identity-blind (confidence) & 0.98 & 1.00 & 1.01 \\
\dotr\ Random order & 1.00 & 1.00 & 1.00 \\
\dotp\ TLC & 1.72 & 1.64 & 1.35 \\
\dotf\ \ffsfaith & 1.00 & 1.00 & 1.01 \\
\dotfair\ \ffsfair & $-0.01$ & $-0.01$ & 0.00 \\
\midrule
\dotp\ Threshold, peeking & \multicolumn{3}{c}{3.39} \\
\dotb\ Threshold, blind & \multicolumn{3}{c}{1.00} \\
\bottomrule
\end{tabular}
\caption{Slope $\beta$ in the synthetic model ($T=1$; 95\% CIs within $\pm0.08$).}
\label{tab:sim}
\end{table}

\paragraph{Order effects.} With consistent conditionals, confidence order and random order (both identity-blind, one token per step) give the same fork distribution at $T=1$ (mean absolute difference 0.005, $p=0.09$ over 19 priors) and different ones at lower temperature (0.005 at $T=0.7$, $p=3\times10^{-4}$; 0.009 at $T=0.5$; 0.019 at $T=0.3$), with confidence order favoring the majority. With two tokens per step at $T=1$ the orders do not differ ($p=0.55$).

\section{Experimental Details}
\label{app:details}

\paragraph{Prompts.} The three templates are ``Write a short story (about 80 words) about a/an \{occupation\}.'', ``Write a short story (about 80 words) describing one day in the life of a/an \{occupation\}.'' and ``Write a brief fictional biography (about 80 words) of a/an \{occupation\}.'', each in the model's chat template without a system prompt. TANGO prompts are ``Continue the following text in 2-3 sentences:'' followed by the TANGO context. Opinion prompts list the question and options A--E and ask for ``the letter of the option you agree with most, followed by a one-sentence reason'' in the format ``<letter>. <reason>''.

\paragraph{Lexicons.} The female group contains she, her, hers, herself; the male group he, him, his, himself; \emph{they} contains they, them, their, theirs, themselves; each in every single-token spelling of the tokenizer (capitalization and leading space). Answer-letter groups contain the letter with and without a leading space or parenthesis.

\paragraph{Decoding.} Stories: 96 tokens, 48 steps. TANGO: 48 tokens, 24 steps, blocks of 24. Opinions: 32 tokens, 16 steps. The default setting ($T=0.5$, top-$p=0.9$) was chosen as the highest temperature at which the identity-blind samplers of both models produce fluent English (judged fluency within 0.5 of the peeking sampler in a held-out set of occupations), a criterion that uses fluency only. In Dream's default decoding, end-of-sequence tokens are proposed with high confidence and committed early, which truncates many stories to a few words; we defer end-of-sequence commits to the last step for every Dream sampler.

\paragraph{Belief audit.} Spearman correlations between the count belief and the faithful pronoun share are 0.968 (\llada) and 0.952 (\dream), with a mean absolute error of 0.049 in both against a sampling floor of 0.032. The belief is 2.6 points more female than the samples in both models ($[1.3, 3.9]$ and $[1.1, 4.2]$; TOST $p=0.0004$ and $0.0014$ for a $\pm5$-point margin). Random and confidence orders differ by $-0.007$ $[-0.043, 0.029]$ and $0.001$ $[-0.039, 0.040]$.

\paragraph{Compute and licenses.} Generation took about 150 GPU-hours and labeling about 40, on 40--48\,GB GPUs. LLaDA-8B-Instruct is under the MIT license, Dream-v0-Instruct-7B and Qwen2.5 under Apache-2.0, Llama-3.3 under the Llama~3.3 Community License, Winogender under MIT, TANGO under CC BY-SA 4.0, and GlobalOpinionQA under CC BY-NC-SA 4.0; our use is non-commercial research.

\section{Label Validation}
\label{app:labels}

The judge (Qwen2.5-32B-Instruct, temperature 0, JSON-constrained decoding) returns the protagonist, the protagonist's gender (female, male, non-binary, unclear, multiple), the pronouns used, whether the story marks the protagonist's gender as unusual, fluency (1--5), coherence (1--5) and instruction following. Rule-based labels count gendered pronouns. Where both assign female or male, they agree on 98.2--99.8\% of stories (mean 99.3\%). About 7\% of stories receive a gendered label from only one source or different ones; on these, a second judge (Llama-3.3-70B-Instruct with 4-bit AWQ quantization, same prompt and output schema) sides with the primary judge in 61\% of cases and with the rules in 32\%. Rule-based labels also count pronouns of secondary characters; this adds noise that flattens steep curves, consistent with the smaller $\Delta\beta$ for \llada under rules (Table~\ref{tab:labels}). 

\paragraph{Human annotation.} Two authors labeled 300 stories independently with the judge's definitions, blind to model, sampler and automatic labels: 200 stratified over eight conditions (25 each from \llada's peeking, identity-blind, random-order, \ffsfair and greedy \ffsfair samplers and \dream's peeking, identity-blind and \ffsfair samplers) and 100 from the disagreement set (\llada's peeking and identity-blind samplers). A third author resolved the 18 disagreements. Agreement on the protagonist's gender is $\kappa=0.90$ (94\%). On the stratified stories the judge is correct in 98.5\% of cases and the rules in 91.0\%; the judge never assigns the opposite gender, and its three errors label a non-binary protagonist as unclear. On the disagreement set the judge is correct in 72\% of cases and the rules in 23\%; 22 of the judge's 28 errors are abstentions (unclear) and one assigns the opposite gender, whereas most rule errors come from stories that also contain pronouns of other characters. Judge errors thus remove stories from the female and male counts rather than flip them. Human fluency ratings correlate with the judge's (Spearman 0.80; 0.70 between annotators) and order the eight conditions as the judge does (0.90 across condition means). Neither annotator marked any story as remarking on the gender as unusual; the judge marked one of the 200 stratified stories.

\begin{table}[h]
\centering\small
\begin{tabular}{@{}lcc@{}}
\toprule
Labels & \llada $\Delta\beta$ & \dream $\Delta\beta$ \\
\midrule
Primary judge & 0.28 $[0.08, 0.51]$ & 0.64 $[0.26, 1.29]$ \\
Second judge & 0.21 $[0.04, 0.42]$ & 0.64 $[0.25, 1.29]$ \\
Pronoun rules & 0.15 $[-0.04, 0.38]$ & 0.61 $[0.24, 1.23]$ \\
\bottomrule
\end{tabular}
\caption{Peeking minus identity-blind slope under three label sources.}
\label{tab:labels}
\end{table}

\section{Additional Amplification Results}
\label{app:amp}

\paragraph{Untruncated $T=1$.} With one token per step the identity-blind rule matches the belief: its slope is 1.01 for \llada and 1.11 for \dream, and random order gives 0.87 and 0.89 (all 95\% CIs contain 1). With two tokens per step both orders flatten the belief slightly (identity-blind 0.76 and 0.72, random 0.77 and 0.73), while peeking stays at 1.73 and 1.87. In either case the text degrades: CJK characters appear in 75--87\% of \llada's identity-blind outputs and 34--44\% of \dream's, against 11\% and 1\% for peeking.

\paragraph{Count belief as the reference.} With the count belief on the horizontal axis, all slopes rise by 9--16\% and their order is unchanged (\llada: peeking 2.10, identity-blind 1.77, random 1.69, \ffsfaith 1.13; \dream: 3.32, 2.61, 2.42).

\paragraph{Robustness of the peeking gap.} With one token per step both \llada rules are equally fluent (4.99 vs.\ 4.98) and $\Delta\beta=0.28$ $[0.09, 0.60]$; with 1, 2 and 4 tokens per step $\Delta\beta$ is 0.28, 0.28 and 0.26. In the default setting the GLM intercepts $\alpha$ (the log-odds of a female protagonist at $\belief=0.5$) are $-0.79$ (peeking), $-0.83$ (identity-blind) and $-0.08$ (random) for \llada, and between $-0.81$ and $-0.47$ for all \dream samplers other than \ffs.

\paragraph{Temperature sweep and parallelism.} Table~\ref{tab:sweep} lists \llada's slopes and intercepts over temperature (untruncated) and over tokens per step (default setting). \ffsfaith has $\beta=1.02$--$1.04$ at every temperature (slope of $\log\beta$ on $\log(1/T)$: 0.011 $[-0.002, 0.026]$). In a \dream sweep on one template (T1, 60 prompts), \ffsfaith's slope is also constant in $T$ but lies at 1.16--1.19; \dream's \ffs slope on this template is 1.17 in the default setting too, against 1.01 and 0.97 on the other two. The slope of $\log\beta$ on $\log(1/T)$ is 0.80 $[0.64, 0.97]$ for the identity-blind rule and 0.19 $[0.08, 0.31]$ for peeking. Between $T=0.3$ and $1.3$ the identity-blind rule commits a token other than the position's mode in 10\% to 90\% of commits and the peeking rule in 5\% to 38\%; $\Delta\beta$ grows from 0.18 at $T=0.3$ to 0.97 at $T=1$. The ordering effect on the intercept (random minus identity-blind) is 0.73 at $T=0.3$, 0.63 at 0.5, 0.47 at 0.7, 0.19 at 1.0 and $-0.08$ at 1.3, and 0.46, 0.76 and 0.69 with 1, 2 and 4 tokens per step.

\begin{table}[h]
\centering\small
\setlength{\tabcolsep}{2.6pt}
\begin{tabular}{@{}l ccc ccc c@{}}
\toprule
 & \multicolumn{3}{c}{$\beta$} & \multicolumn{3}{c}{$\alpha$} & Flu. \\
\cmidrule(lr){2-4}\cmidrule(lr){5-7}
 & Peek & Blind & Rand. & Peek & Blind & Rand. & Blind \\
\midrule
$T=0.3$ & 2.00 & 1.81 & 1.82 & $-0.99$ & $-0.94$ & $-0.21$ & 3.38 \\
$T=0.5$ & 1.80 & 1.44 & 1.49 & $-0.90$ & $-0.71$ & $-0.08$ & 3.16 \\
$T=0.7$ & 1.78 & 1.16 & 1.18 & $-0.72$ & $-0.64$ & $-0.18$ & 2.75 \\
$T=1.0$ & 1.73 & 0.76 & 0.77 & $-0.33$ & $-0.38$ & $-0.19$ & 1.72 \\
$T=1.3$ & 1.43 & 0.57 & 0.55 & $-0.09$ & $+0.32$ & $+0.24$ & 1.01 \\
\midrule
$k=1$ & 2.08 & 1.80 & 1.69 & $-0.45$ & $-0.28$ & $+0.17$ & 4.98 \\
$k=2$ & 1.80 & 1.53 & 1.49 & $-0.79$ & $-0.83$ & $-0.08$ & 3.16 \\
$k=4$ & 1.66 & 1.39 & 1.52 & $-1.27$ & $-1.11$ & $-0.43$ & 1.71 \\
\bottomrule
\end{tabular}
\caption{\llada slope and intercept over temperature and tokens per step $k$.}
\label{tab:sweep}
\end{table}

\paragraph{Third model.} LLaDA-1.5 \citep{zhu2025llada15} has slopes of 1.83 (peeking), 1.60 (identity-blind) and 1.46 (random), $\Delta\beta=0.23$ $[0.08, 0.44]$; intercepts of $-0.66$ $[-1.19, -0.26]$, $-0.53$ $[-0.92, -0.20]$ and $-0.03$ $[-0.25, 0.18]$; \ffsfaith $\beta=0.99$ and \ffsfair target error 0.001. Its one-pass belief matches faithful samples (Spearman 0.90) and is close to LLaDA's (Spearman 0.996 across prompts).

\paragraph{AR reference.} Qwen2.5-7B-Instruct writes female protagonists in 0.786 of stories at $T=1$, 0.793 at $T=0.7$, 0.805 at $T=0.5$ with top-$p=0.9$, and 0.817 under greedy decoding; with the gender specified in the prompt, it complies in 99.9\% of stories and marks 0.6\%.

\section{Mechanism Analyses}
\label{app:mech}

\paragraph{Commit-rate ratio.} We record, for every step up to the fork, each masked position whose proposal is a female or male pronoun, whether it was committed, its $\max_vp_i(v)$, and its group masses. The majority proposal at a position is the group with the larger mass there. Within five bins of $\max_vp_i(v)$ we compute the commit rates of majority and minority proposals and combine them with weights proportional to the number of minority proposals; the ratio for peeking divided by that for the identity-blind rule estimates $\Gam$. CIs bootstrap over prompts. Data: 60 prompts of one template, 20 stories each. The ratios are 2.03 (peeking) and 0.89 (identity-blind) for \llada and 62.5 and 1.18 for \dream, giving $\Gam=2.28$ $[1.84, 2.87]$ and 53 $[21, 185]$; with one token per step \llada's $\Gam$ is 2.63 $[1.91, 4.10]$.

\paragraph{Multi-slot competition.} When several masked positions propose gendered pronouns in the same step and only one of them is committed, a peeking rule commits the most probable proposal, so the fork's value is decided by a competition among slots. At such steps the committed proposal belongs to the group that holds the larger pronoun mass at that step in 81\% of cases under \llada's peeking rule and 58\% under its identity-blind rule (base rate 62\%; difference $+0.23$ $[0.12, 0.34]$), and in 98\% versus 79\% in \dream (base rate 77\%; $+0.19$ $[0.13, 0.25]$). Defining the majority by the step-0 belief instead gives no difference for \llada ($+0.02$), because the belief drifts before the fork: the two definitions agree on 66--68\% of \llada's fork steps and 83--86\% of \dream's, and 84--92\% of the disagreements are prompts whose step-0 belief is female but whose pronoun mass at the fork is male.

\paragraph{Fork state.} The first gendered word determines the protagonist's gender in 97--99.7\% of stories. Table~\ref{tab:fork} lists how often the fork position's female share is already below 0.1 or above 0.9 when the first pronoun is committed, and the drift of that share on belief-balanced prompts.

\begin{table}[h]
\centering\small
\setlength{\tabcolsep}{4pt}
\begin{tabular}{@{}llcc@{}}
\toprule
Model & Sampler & Decided & Drift (balanced) \\
\midrule
\llada & Peeking & 65\% & $-0.22$ $[-0.30, -0.12]$ \\
\llada & Identity-blind & 65\% & $-0.20$ $[-0.27, -0.12]$ \\
\llada & Random order & 40\% & $-0.04$ $[-0.07, -0.01]$ \\
\dream & Peeking & 77\% & $-0.28$ $[-0.41, -0.15]$ \\
\dream & Identity-blind & 80\% & $-0.26$ $[-0.37, -0.13]$ \\
\dream & Random order & 70\% & $-0.25$ $[-0.35, -0.17]$ \\
\bottomrule
\end{tabular}
\caption{Fork position already decided at the first pronoun, and its pre-fork drift.}
\label{tab:fork}
\end{table}

\paragraph{Context transplant.} We generate the 60 prompts of one template with confidence order and with random order (identity-blind, default setting), save the canvas just before the first gendered word is committed, and finish every saved canvas with each order using fresh randomness. On belief-balanced prompts the male share is 0.72 (confidence context, confidence continuation), 0.69 (confidence, random), 0.60 (random, confidence) and 0.49 (random, random). The context effect under random continuation is $+0.19$ $[0.07, 0.29]$; the continuation effect is $+0.11$ $[0.03, 0.19]$ after a random-order context and $+0.03$ $[-0.02, 0.08]$ after a confidence-order context.

\paragraph{Ordering intervention.} Excluding the gendered pronouns (and, in a second variant, gendered titles and nouns such as ``Mr'' and ``woman'') from the identity-blind ordering score changes the intercept by $+0.11$ $[-0.02, 0.26]$ and $+0.13$ $[-0.00, 0.27]$ in \llada and by $-0.16$ $[-0.34, -0.00]$ and $-0.12$ $[-0.30, 0.04]$ in \dream, where the pronoun is committed much later (step 25 instead of 11) and the drift has more time to act. On belief-balanced prompts, confidence ordering also commits gendered titles and nouns before the first pronoun more often than random order (10\% and 13\% of \llada's stories vs.\ 4\%, almost always male), and stories with such an early title end with that gender in 88--94\% of cases.

\section{Fork-First Sampling Details}
\label{app:ffs}

\paragraph{Intervention rate and consistency.} \ffs intervened in every occupational prompt. Protagonist consistency is 0.99 (\llada) and 0.98 (\dream). The pooled female share of \ffsfair by belief decile deviates from 0.5 by at most 0.05 for \llada and 0.07 for \dream, in the direction of the belief at the extremes.

\paragraph{Nominative planting.} An earlier variant planted the group's most probable form of any case at $i^\star$, often \emph{her}, which at extreme beliefs sometimes binds to a secondary character. With the same draws of $g$, planting the nominative form at $j^\star$ raises consistency on plants against the belief from 0.91 to 0.96 in \dream (difference $+0.05$ $[0.03, 0.08]$; \llada $+0.01$ $[-0.00, 0.02]$), halves the largest decile deviation of \ffsfair in \dream (0.14 to 0.07), and leaves fluency and marked narratives unchanged.

\paragraph{Faster decoding.} With four tokens per step \ffsfaith has $\beta=1.02$ (\llada) and 1.06 (\dream) and \ffsfair a target error of 0.008 and 0.013; under Fast-dLLM's threshold rule ($\lambda=0.9$) the values are 1.02 and 0.007. Every \ffs intercept has a 95\% CI containing 0, and the male share on belief-balanced prompts is 0.51--0.52.

\paragraph{Efficiency.} The belief read-out is one forward pass on the prompt and a fully masked canvas, 2\% of the cost of a 48-step generation.

\paragraph{Greedy decoding.} With greedy decoding every prompt yields two stories under \ffsfair, one per planted gender, so diversity is that of the draw.

\paragraph{Distinguishability AUC.} For each model, method and protagonist gender, we take stories labeled with that gender, match their number per prompt to the default sampler's stories with the same gender, remove gendered pronouns, titles and nouns, and represent each story by the presence of unigrams and bigrams (20,000 most frequent). An L2-regularized logistic regression is trained with 5-fold cross-validation grouped by prompt; we report the out-of-fold AUC averaged over three matching seeds. The default sampler is the reference and has no AUC; greedy decoding writes one story per prompt and gender, too few to train the classifier. Changing only the sampler also shifts word use (random order: .86 and .93), so the comparison that isolates an intervention is among methods that decode with the default sampler, as \ffs and the prompt-based fixes do. Words that most distinguish prompt-specified stories from the default's describe the character rather than a plot (``very'', ``good'', ``person'', ``loved'', ``married'') or a daily routine (``wakes'', ``dressed'', ``breakfast''); \ffs stories do not show such a shift.

\section{Opinions and Misgendering}
\label{app:beyond}

\paragraph{Opinions.} GlobalOpinionQA has 2,556 questions; after removing refusal options (``Refused'', ``DK'', ``Don't know'', ``No answer''), 2,370 have two to five options, from which we sample 250 stratified by source and number of options. The belief is read at the position of the fully masked canvas with the largest answer-letter mass, which is always the first generated position, where letters hold a median of 97\% (\llada) and 91\% (\dream) of the probability mass. The floor simulates 200 multinomial draws from the belief with the same number of parsed answers. In the default setting ($T=0.5$, top-$p=0.9$) the lower temperature itself collapses the answers (collapsed questions: \llada 37\% peeking and 29\% identity-blind; \dream 42\% and 35\%), peeking adds less (entropy-ratio difference $-0.13$ and $-0.06$), and \ffsfaith stays at the floor (TV 0.111 and 0.102).

\begin{table}[h]
\centering
\small
\setlength{\tabcolsep}{3.2pt}
\begin{tabular}{@{}l rrr rrr@{}}
\toprule
 & \multicolumn{3}{c}{\llada} & \multicolumn{3}{c}{\dream} \\
\cmidrule(lr){2-4}\cmidrule(l){5-7}
 & $H$ ratio & TV & Coll. & $H$ ratio & TV & Coll. \\
\midrule
\dotp\ Peeking & 0.42 & .252 & 13.6 & 0.45 & .203 & 16.0 \\
\dotb\ Identity-blind & 0.78 & .189 & 2.8 & 0.87 & .145 & 2.0 \\
\dotf\ \ffsfaith & 0.93 & .111 & 0.8 & 0.91 & .102 & 0.4 \\
Sampling floor & 0.91 & .107 & -- & 0.91 & .101 & -- \\
\bottomrule
\end{tabular}
\caption{Opinion distributions ($T=1$). Floor: exact sampling from the belief.}
\label{tab:beyond}
\end{table}

\paragraph{Declared-pronoun planting.} Planting the nominative form of the declared family at the first-block position with the largest third-person-pronoun mass (\emph{fae} as its two subtokens), then decoding with the default sampler, changes the misgendering rate of the remaining pronouns by $-0.37$ $[-0.53, -0.21]$ (xe), $-0.13$ $[-0.23, -0.05]$ (ey) and $-0.18$ $[-0.35, -0.03]$ (fae) in \llada and by $-0.35$, $-0.34$ and $-0.17$ in \dream (all CIs exclude 0). It lowers misgendering for \emph{he} and \emph{they} in both models; for \emph{she} it is unchanged in \llada and rises slightly in \dream, from 1.8\% to 3.7\% ($+0.019$ $[-0.000, 0.045]$).

\paragraph{Misgendering.} Table~\ref{tab:tango} gives misgendering rates over all third-person pronouns by family. At untruncated $T=1$, \llada's peeking rule misgenders 5 points less than its identity-blind version, whose text degrades at that temperature.

\begin{table}[h]
\centering\small
\setlength{\tabcolsep}{3.5pt}
\begin{tabular}{@{}ll cccccc@{}}
\toprule
 & & she & he & they & xe & ey & fae \\
\midrule
\llada & Belief & .05 & .17 & .22 & .96 & .92 & -- \\
 & Peeking & .04 & .12 & .17 & .95 & .93 & .79 \\
 & Blind & .05 & .15 & .17 & .96 & .95 & .79 \\
 & Random & .05 & .12 & .22 & .96 & .96 & .80 \\
\midrule
\dream & Belief & .04 & .14 & .31 & .89 & .71 & -- \\
 & Peeking & .02 & .15 & .29 & .57 & .72 & .89 \\
 & Blind & .02 & .15 & .29 & .62 & .72 & .91 \\
 & Random & .03 & .14 & .30 & .72 & .77 & .90 \\
\bottomrule
\end{tabular}
\caption{TANGO misgendering rate by declared pronoun family.}
\label{tab:tango}
\end{table}

\section{Example Generations}
\label{app:examples}

LLaDA, prompt ``Write a short story (about 80 words) about a mechanic.'', target gender female.

\paragraph{\ffsfair (planted ``she'').} ``There was once a mechanic named Sarah who had a knack for fixing cars. One day, while working on an old 76 Fordala Sr., she discovered a hidden compartment in the engine. Inside, she found a dusty, vintage car with a key. [...]''

\paragraph{Prompt-specified (``The main character is a woman.'').} ``She was was a mechanic mechanic. She was a woman mechanic. She was a woman mechanic. She was a woman mechanic. [...]'' (judged marked, fluency 2).

LLaDA, prompt about an engineer, target female. \ffsfair (planted ``She''): ``Lena, a brilliant engineer, dreamed of designing a bridge that would connect two distant towns. She worked tirelessly, pouring over blueprints and calculations. [...]'' Prompt-specified: ``Lena was a brilliant engineer who had always dreamed fascinated by the world of technology. She spent her days in in her lab, tinkering with gadgets and designing new machines. [...]''

\end{document}